# Improve Machine Learning carbon footprint using Parquet dataset format and Mixed Precision training for regression models

# Part II

**Andrew Antonopoulos**

andrew.antonopoulos@sony.com

## Abstract

This is the 2nd part of my dissertation for my master's degree and compared the power consumption using the Comma-Separated-Values (CSV) dataset format and Parquet dataset format with the default floating point (32-bit) and Nvidia's mixed precision (16-bit and 32-bit) while training a regression ML model. The same custom PC as per the 1st part [1] was built to perform the experiments, and different ML hyper-parameters, such as batch size, neurons, and epochs, were chosen to build Deep Neural Networks (DNN). A benchmarking test with default hyper-parameter values for the DNN was used as a reference, while the experiments used a combination of different settings. The results were recorded in Excel, and descriptive statistics were chosen to calculate the mean between the groups and compare them using graphs and tables. The outcome was positive when using mixed precision combined with specific hyper-parameters. Compared to the benchmarking, the optimisation for the regression models reduced the power consumption between 7 and 11 Watts. The regression results show that while mixed precision can help improve power consumption, we must carefully consider the hyper-parameters. A high number of batch sizes and neurons will negatively affect power consumption. However, this research required inferential statistics, specifically ANOVA and T-test, to compare the relationship between the means. The results reported no statistical significance between the means in the regression tests and accepted $H_0$. Therefore, choosing different ML techniques and the Parquet dataset format will not improve the computational power consumption and the overall ML carbon footprint. However, a more extensive implementation with a cluster of GPUs can increase the sample size significantly, as it is an essential factor and can change the outcome of the statistical analysis.

## 1  Introduction

The greenhouse effect is a natural phenomenon related to the sun's radiation, which travels towards the Earth [2]. The radiation reaches the earth and is absorbed by the land and ocean, and some are released toward space [2]. Most of it is captured and retained by greenhouse gases, a combination of chemical compounds that help keep Earth at a suitable temperature for all living beings [3]. Gases like carbon dioxide are produced naturally or by human activities, and by increasing it will also increase the Earth's temperature, affecting everyone's life [3]. The carbon footprint is the total amount of carbon dioxide emitted by human actions and is measured in grams of $CO_2$ (Carbon dioxide) equivalent per kilowatt hour (gCO2e/kWh) [4]. The higher the carbon footprint, the more impact it will have on the environment.

Machine Learning (ML) has become very popular in many industries, and various services, such as cybersecurity, healthcare, and finance, have adopted it [5]. Millions of people use ML services hosted in the Cloud and specifically in big data centres [6]. This forces service providers to build big data centres to store the hardware and support growth. The data centres require cooling systems and power generators to maintain thousands of servers, consuming substantial power sources such as water and electricity [6]. Therefore, ML services are increasing and overloading many data centres worldwide, which can affect their sustainability, eventually increasing the carbon footprint and affecting the environment.

Data centres are using energy from non-fossil-fuelled technologies (solar, wind, hydro) instead of fossil-fuelled technologies (coal, oil, gas) [4]. However, there are no carbon-free forms of generating energy [4], and optimising ML services is a potential candidate to help reduce the carbon footprint.

## 2  Background

Table 1 shows the most common row and column-based data stores [7]. The row-based stores data as a list of rows, and the column-based stores data as a list of columns.

| Types | Row-Based (Text) | Column-Based (Binary) |
|---|---|---|
| Parquet | | √ |
| ORC | | √ |
| CSV | √ | |
| JSON | √ | |
| XML | √ | |

**Table 1: Row and column-based file formats**

There is a preference for using CSV, JSON, or XML files, mainly for regression models, because they are human-readable and can be used easily with ML frameworks [8]. While extra work is required to read other formats [9], Pandas, a Python data analysis library, can also use Parquet to load and store datasets [8].

Furthermore, Parquet has become the most popular file format for Pandas because of the wide variety of encoding algorithms [10] and the broad support from big data frameworks and large-scale query providers [11].

Additionally, Parquet is more optimised for processing structured data against column-based file types, such as JSON, and could achieve better speed for reading data [12]. Generally, column-based datasets use less storage because of their aggressive dictionary encoding, but they are binary files and cannot be read [13]. Also, column-based datasets are not designed to be GPU-friendly, which can affect ML implementations [40]. Nevertheless, companies such as Twitter and Netflix, which store large amounts of data, use Parquet files because they allow more data to be held in the same physical space [41].

Figure 1 [15] shows the row-based storage layout, where data is stored row by row. This format is suitable for small datasets or scenarios where the entire row of data is frequently accessed or modified. In a column-based storage layout, data is stored by columns. This format is ideal for scenarios where analytical queries typically access only a subset of columns, as it allows for more efficient data retrieval and processing.

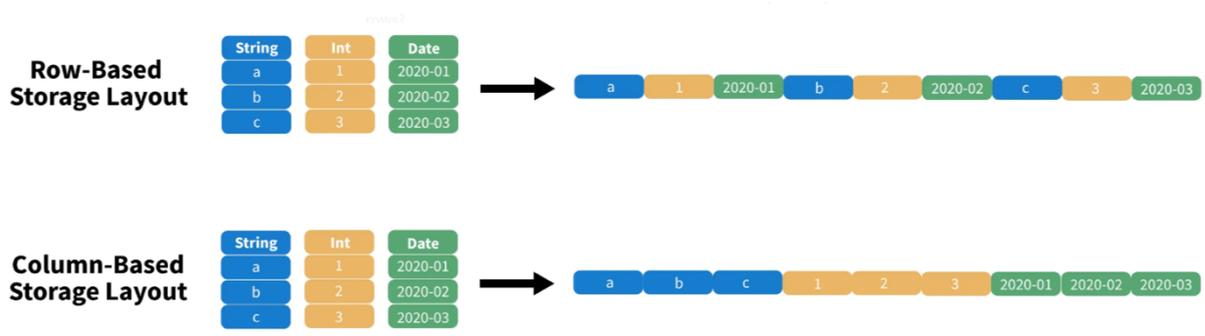

**Figure 1: Research steps during experiments and data collection**

# 3  Methodology

Large datasets with a minimum of 1 GB of data have been used to assess power consumption over a period of time. This is essential because the longer the model training takes, the more power consumption data will be generated. Therefore, the regression CSV data were pre-processed before being used for the model training, ensuring the accuracy of the data.

The regression dataset contains information on used car sales, such as models, prices, and production years, and is in CSV format. The owner published the data on the Kaggle platform [16], which was scraped using web crawlers; it contains most of Craigslist's relevant information on car sales, including columns like price, condition, manufacturer, latitude/longitude, and 18 other categories.

However, CSV and Parquet were used to collect data for analysis and comparison during the experiments.

Figure 2 shows the steps followed to generate and collect data. Various experiments were created by utilising different ML optimisation techniques and hyperparameters. The data were collected into an Excel file and used for analysis during the experiments. This procedure was repeated until it satisfied all the experiment use cases.

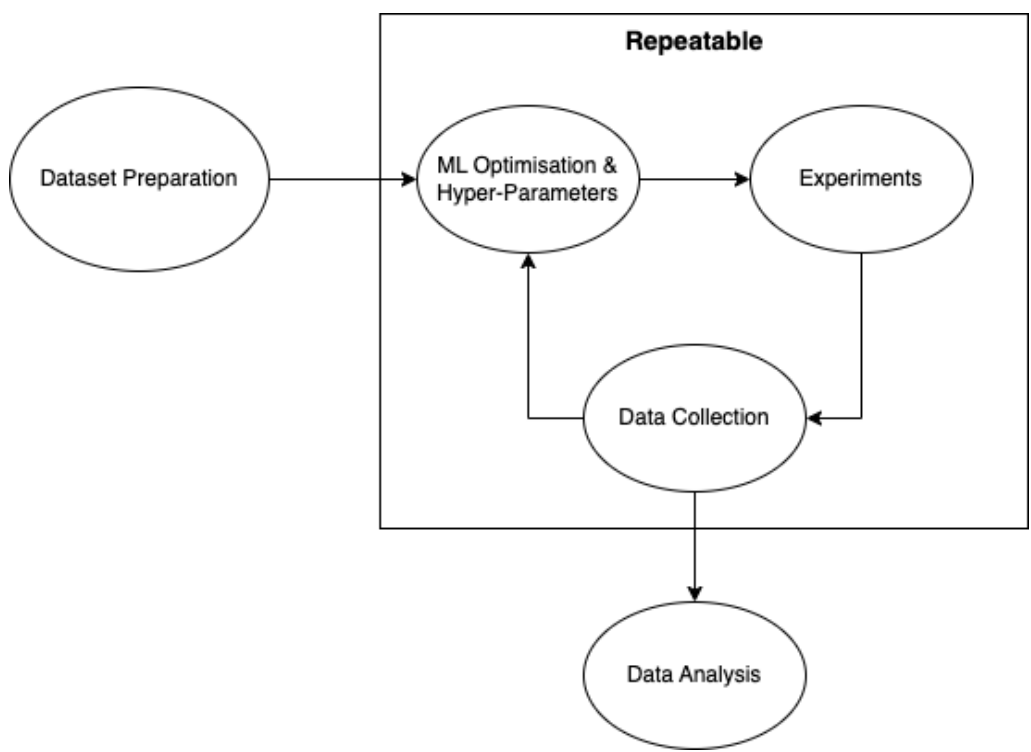

**Figure 2: Research steps during experiments and data collection**

Similar to the 1st part [1], a custom PC was built, which was used during the experiments to produce and collect the data. The hardware components were the following:

| Component (Hardware/Software) | Model |
| --- | --- |
| Motherboard | MSI Z690 DDR4 |
| CPU | Intel i5-12600 |
| Memory | Kingston Fury 32GB (4x8GB) |
| GPU | MSI NVIDIA RTX 4060 16GB GDDR6 (18Gbps/128bit) |
| SSD | Kingston A400 500GB |
| PSU | EVGA 1600w P2 |
| OS | Windows 10 Pro |

Additionally, the same ML framework, TensorFlow and Keras, was used. The main reason was that TensorFlow utilises the GPU more efficiently [17].

Besides GPU utilisation and accuracy, TensorFlow has better memory management than other frameworks, which is essential for large batch sizes and can improve power consumption [18].

TensorFlow requires experience; however, Keras, a high-level API that runs on top of TensorFlow, provides a quick implementation, has a simple architecture, and focuses on the user experience to accelerate the development of DNNs [19]. Therefore, TensorFlow and Keras were used to develop the ML models and perform the experiments.

## 3.1 Collecting computation power consumption data

Identifying the hardware and software to collect power consumption data is a crucial step. The Graphics Processing Unit (GPU) accounts for around 70% of power consumption. In comparison, the Central Processing Unit (CPU) is responsible for 15%, Random Access Memory (RAM) for 10%, and the remaining 5% from other PC components [20]. Therefore, the GPU, CPU and RAM are critical components because they directly impact the ML lifecycle. SSD or HDD are also crucial but are used by the operating system and other processes, so it is challenging to clarify the direct relationship to the ML process [21].

Additionally, the same software as the 1st part [1] was used during the experiments to collect the power consumption data in Watts from the GPU, CPU, and RAM and manually from a wattmeter connected to the wall.

1. **Comet** automatically creates an Emissions Tracker object from the code carbon package to visualise the experiment's carbon footprint.
2. **Code Carbon v3.35.3** is lightweight software that seamlessly integrates into the Python codebase. It estimates the amount of carbon dioxide ($CO_2$) that the personal computing resources produce when executing the code.
3. **HWiNFO v7.66-5271**, focuses on hardware and categorises all the information it collects into sections. It can also collect power consumption for the CPU and GPU.
4. **Core Temp v1.18.1**, is a compact and powerful program for monitoring processor temperature and other vital information, such as power consumption.
5. **MSI Afterburner v4.6.5**, provides an on-screen display, hardware monitoring, custom fan profiles, and video capture. Additionally, it includes power consumption for the GPU and CPU.
6. **Corsair iCUE v5.9.105,** allows customisation of its various supported components and peripherals and provides information on how the GPU and CPU are used.
7. **Intel Power Gadget v3.6** is a software-based power estimation tool explicitly designed to monitor power consumption and utilisation for Intel Core processors.
8. **Wattmeter** was used to monitor the overall power consumption connected to the wall socket and the PC's power supply directly to the wattmeter.

## 3.2 ML optimisation techniques

Optimisation is crucial when creating a more efficient DNN because it has a certain level of complexity. Hyper-parameter optimisation techniques, such as the number of hidden layers, batch size, neurons, and epochs, cannot be modified individually and manually because they require a lot of time and experience [22]. If a non-optimal hyper-parameter is chosen for a particular reason, the DNN will consume more processing power [23]. The hyper-parameter will require fine-tuning to achieve the ideal results, but DNNs may fail to train or receive inefficient results because of the non-optimal values [24].

As per the classification test in 1st part [1], the same hyper-parameters and mixed precision were used for the benchmarking and experiments, as shown below:

- **Neurons** determine the amount of information stored in the network, and more neurons allow us to learn more complex patterns. It can also increase the number of network connections, which requires more computational resources [25].
- **Batch size** is the number of training samples used to train a neural network. To fully take advantage of the GPU's processing, the batch size should be a power of 2 [26].
- **Epochs** are the number of complete passes of the training dataset through the algorithm's learning process, and the default values were identified during the pre-tests [22].

### 3.3 Power Consumption Data

Figure 3 shows the architecture and how data were collected. Multiple third-party software extracted the RAM, CPU, and GPU utilisation and power consumption data in Watts. The data were collected in an Excel file for comparison and generating the average value. The PSU was connected directly to the wattmeter, but reading the values manually was required because the software was unavailable.

Code Carbon, a Python library, was integrated into the Python code, and data was seamlessly collected while the code was running. However, Code Carbon cannot store historical data, and Comet has been used to retrieve the average value over a period of time. Comet is a web service that pulls data from Code Carbon via an API to monitor GPU and CPU power consumption and utilisation. The collected data from all the software and the wattmeter was imported into Excel for further analysis.

Watts have been chosen because they measure the power consumed by a device. The higher the wattage, the more significant the amount of electrical power the PC uses over a period of time.

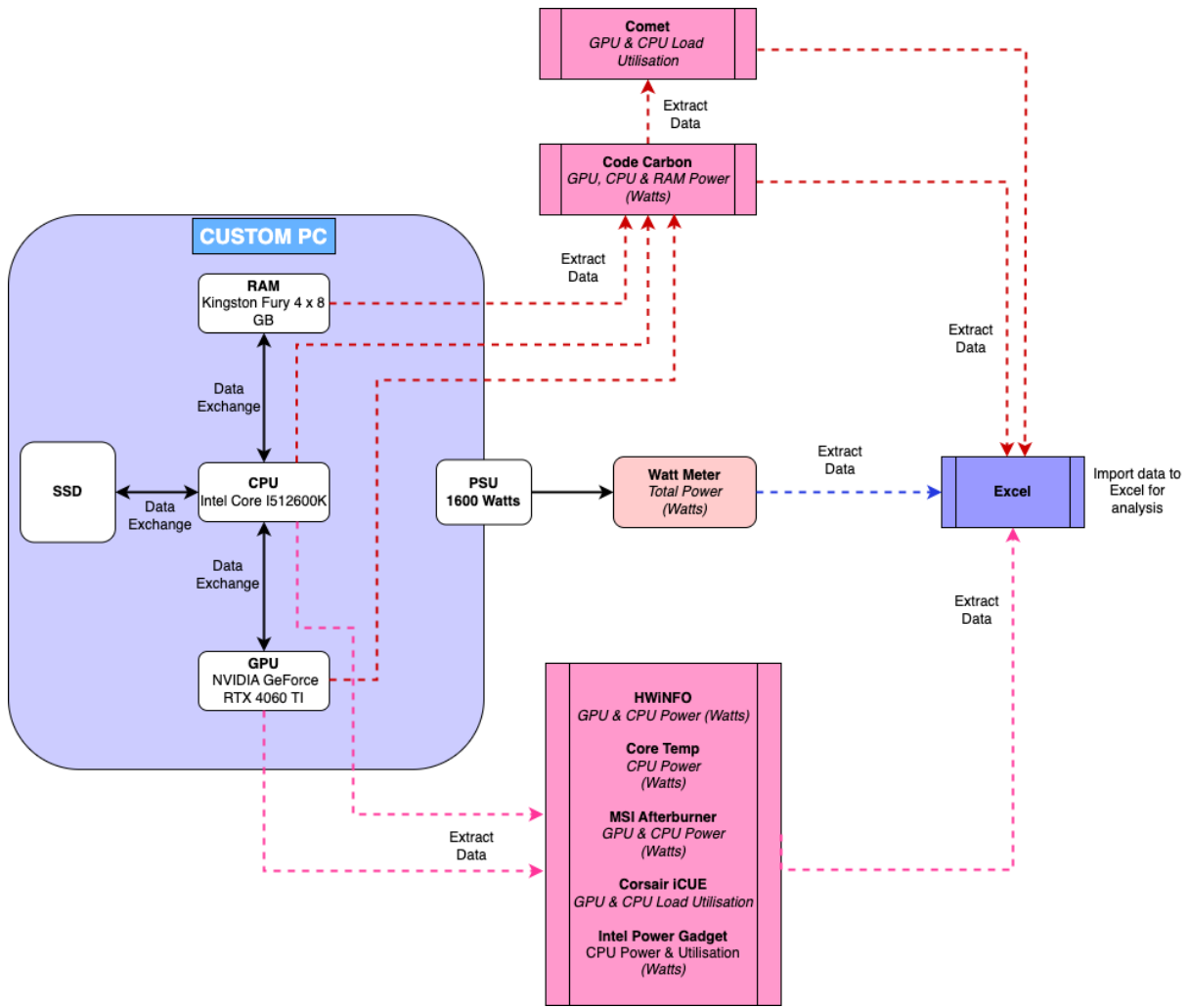

Figure 3: Overall architecture to collect power consumption data

## 3.4 Data Analysis Technique

A similar approach as per the 1st part [1] was used, and descriptive statistics were adapted to assess the central tendency of the power consumption values. The author used a component bar chart to illustrate the comparison between the average of each piece of hardware [27]. However, further analysis of the findings using inferential statistics was required because the differences between the average values were too close. To achieve this task, ANOVA was used to evaluate the relationship between the tests and multiple T-tests were used to check whether the difference between experiments was statistically significant [28]. Figure 4 summarises the steps that followed during the analysis.

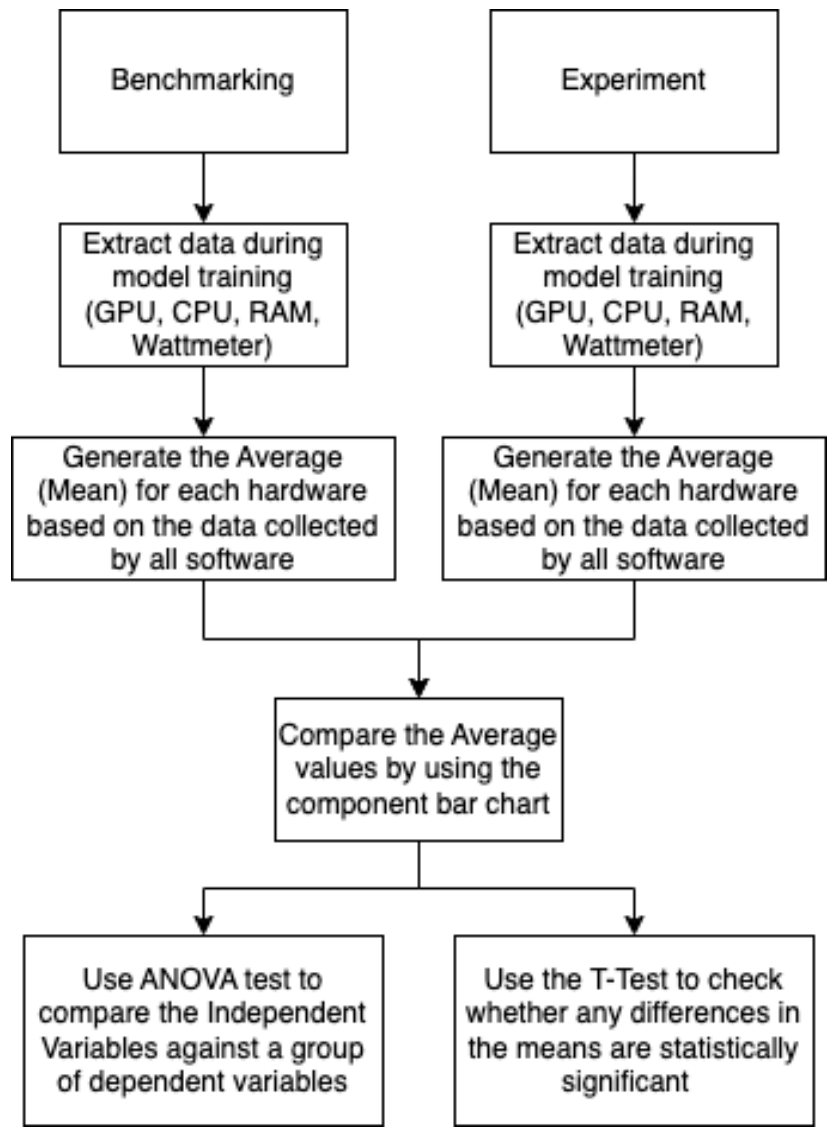

Figure 4: Steps that followed during the data analysis

# 4 Testing and Results

## 4.1 Introduction

The GPU has played a vital role in ML and model training because it is powered by Tensor Cores, which are specialised cores that enable mixed precision and can accelerate training and learning performance [29]. Using a GPU that supports Tensor Cores, we can utilise the mixed precision functionality, accelerating the throughput and reducing AI training times [29].

Therefore, the same GPU, NVIDIA RTX 4060 Ti Ventus, which supports overclocking and operates at 2595 MHz instead of 2565 MHz in standard mode, were used similarly to the 1st part [1]. This frequency indicates how much data it can process per clock cycle. Additionally, it supports the 4th generation of NVIDIA's Tensor Cores and the latest technology in high-performance memory GDDR6 with a capacity of 16 GB. However, the most important is the high memory bandwidth of 18 Gbps, which allows fast data transfer between the GPU memory and the computation cores.

To use the mixed precision, the libraries have been imported into the Python code and configured to be used with the public policy. After the implementation and execution of the code, the mixed precision library checked the GPU and reported the version of the computation capability. The computation capability identifies the features supported by the GPU hardware and is used by applications at runtime to determine which hardware instructions are available [30]. According to the mixed precision Python library, the compute capability version must be more than 7.0. The GPU that has been used for this research have a compute capability version of 8.9, as Figure 5 shows:

```
INFO:tensorflow:Mixed precision compatibility check (mixed_float16): OK
Your GPU will likely run quickly with dtype policy mixed_float16 as it has comp
ute capability of at least 7.0. Your GPU: NVIDIA GeForce RTX 4060 Ti, compute c
apability 8.9
Compute dtype: float16
Variable dtype: float32
```

**Figure 5: Mixed precision and compute capability reported by the Python library**

The above output indicates that the current GPU will use floating-point 16-bit for computations to improve performance and 32-bit for the variables, mainly for numerical stability, so the model trains with the same quality.

During the regression model training benchmarking, the default floating point of 32 bits was used, while all the experiments used only mixed precision.

## 4.2 Regression

Similar to the classification test [1], the initial step was to load the dataset, and the mean type from the descriptive statistics was used to calculate the average.

The original dataset has 426,881 rows and was required to execute pre-tests to determine if the PC's RAM can handle the dataset size during pre-processing. The author tried with different datasets, reducing the rows by 50,000 in each test. The conclusion was that with a dataset of 150,000 rows, the PC's RAM could process the dataset and had enough memory for other processes related to the operating system.

Different Python codes were created, and steps were followed for the CSV and Parquet datasets. However, the data collection procedure was similar, as shown in [Figure 6](#).

The CSV was preprocessed and cleaned before the model training, and specific hyperparameters were used to create the DNNs. To use the Parquet dataset, the author had to load the CSV, convert it to Parquet, and save it to the disk. After this step, the author could reload it into the Python code and use it for the preprocessing. With this procedure, it could be guaranteed that only the Parquet dataset was used during the model training.

After cleaning the dataset, mixed precision and hyper-parameters were chosen, and data were collected before and during the model training, similar to the classification testing [1].

Each test was executed for CSV and Parquet, and the same DNN configuration was chosen to compare the results and identify potential differences.

However, the same issue with the RAM power consumption was applied to the regression tests, and the fixed value from Code Carbon was chosen. Furthermore, because of the dataset's numerical values, the model training took less time and used fewer resources. Additionally, the overall power consumption was taken from the wattmeter.

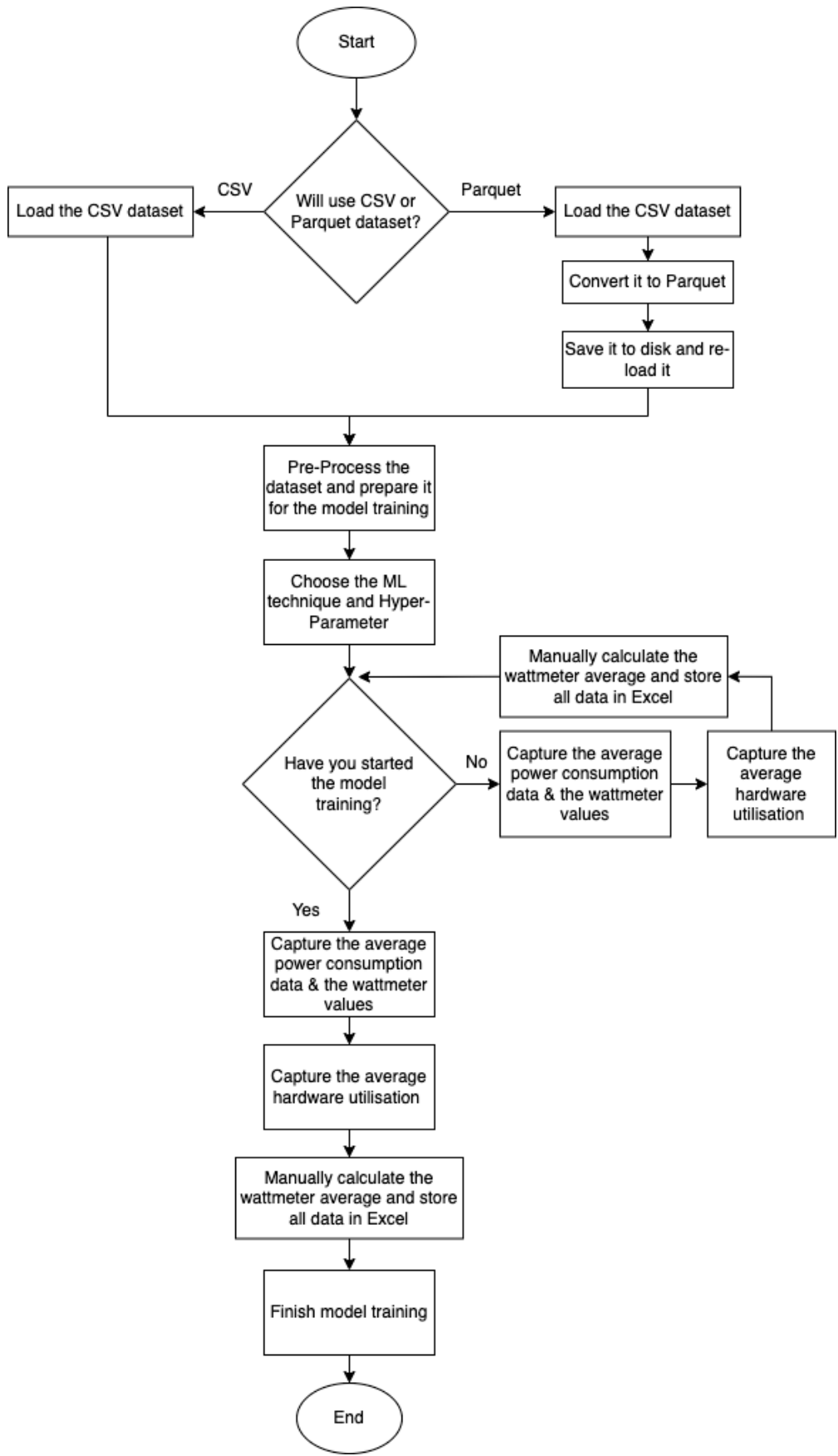

**Figure 6: Flow of the regression testing and data collection**

## 4.3 Benchmarking

Two benchmarking tests have been completed, one for the CSV and another for the Parquet dataset format. Table 2 shows the configuration for the DNNs.

|  | CSV – Benchmarking | Parquet – Benchmarking |
|---|---|---|
| Floating Point | 32 | 32 |
| Batch Size | 256 | 256 |
| Neurons | 1024 | 1024 |
| Epochs | 1500 | 1500 |

**Table 2: Hyper-parameters for the regression benchmarking**

The same methodology has been followed as the classification benchmarking [1]. The floating points were 32 bits, the default value, and neurons were 1024. The regression dataset has numerical values, making it easier for the GPU to process the data. Therefore, the batch size and epochs were adjusted, which is responsible for the duration it takes to train the model. By increasing the epochs, the model training took longer, allowing more accurate measurements to be collected.

Figure 7 shows the power consumption during the CSV benchmarking testing. Before the model training, the power consumption was within normal PC operational values, and the RAM was 12 Watts because of the fixed value from Code Carbon. During model training, the GPU increased to 44 Watts, the CPU to 27 Watts and the overall power consumption to 126 Watts. Figure 8 confirms that GPUs were used during the model training with utilisation at 40%.

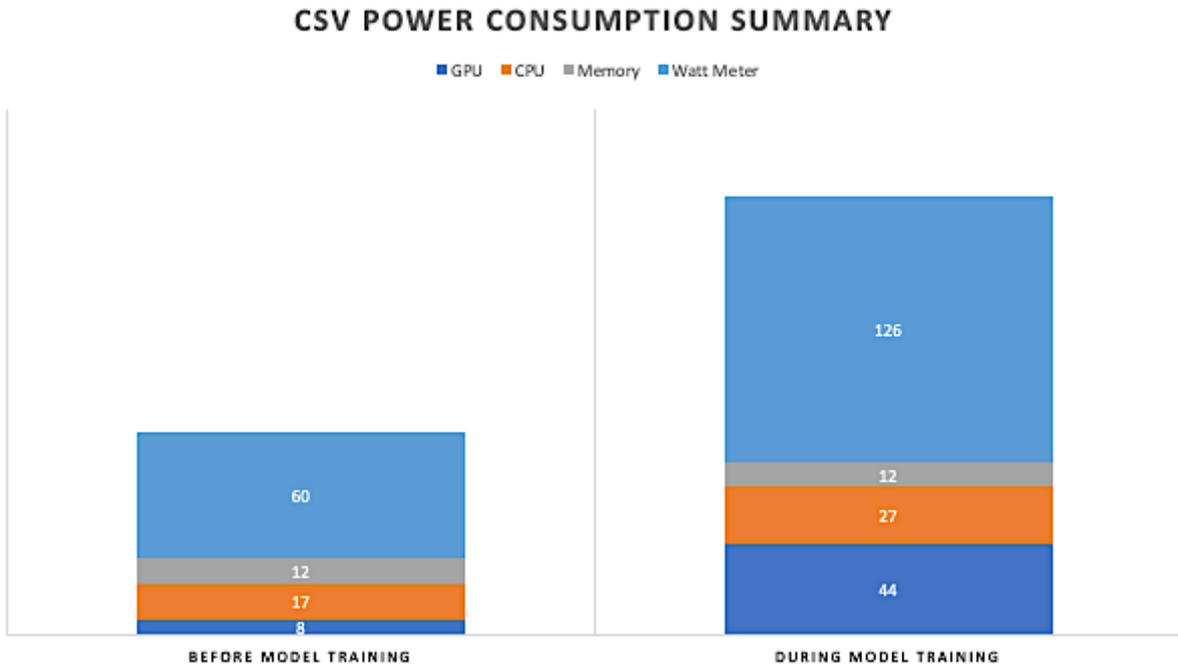

**Figure 7: Power consumption data for the CSV regression benchmarking**

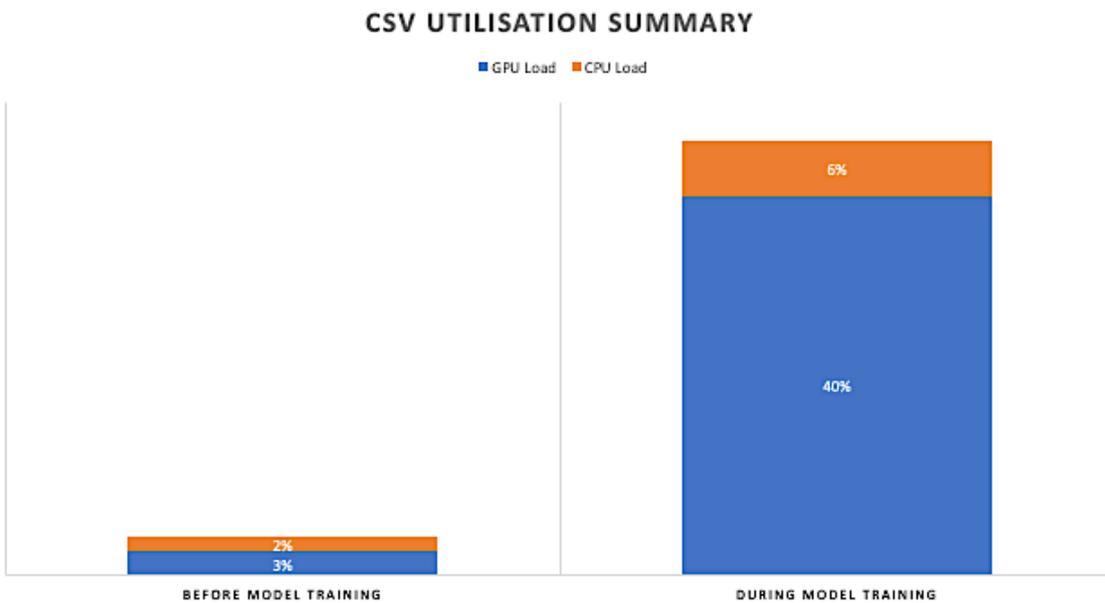

**Figure 8: Hardware utilisation data for the CSV regression benchmarking**

Similarly, Figure 9 shows the power consumption using the Parquet dataset format. Before the model training, the overall power consumption was 61 Watts, GPU 7 Watts and CPU 15 Watts, but during the model training, the values increased to 126 Watts for the overall power consumption, 44 for the GPU, and 27 for the CPU.

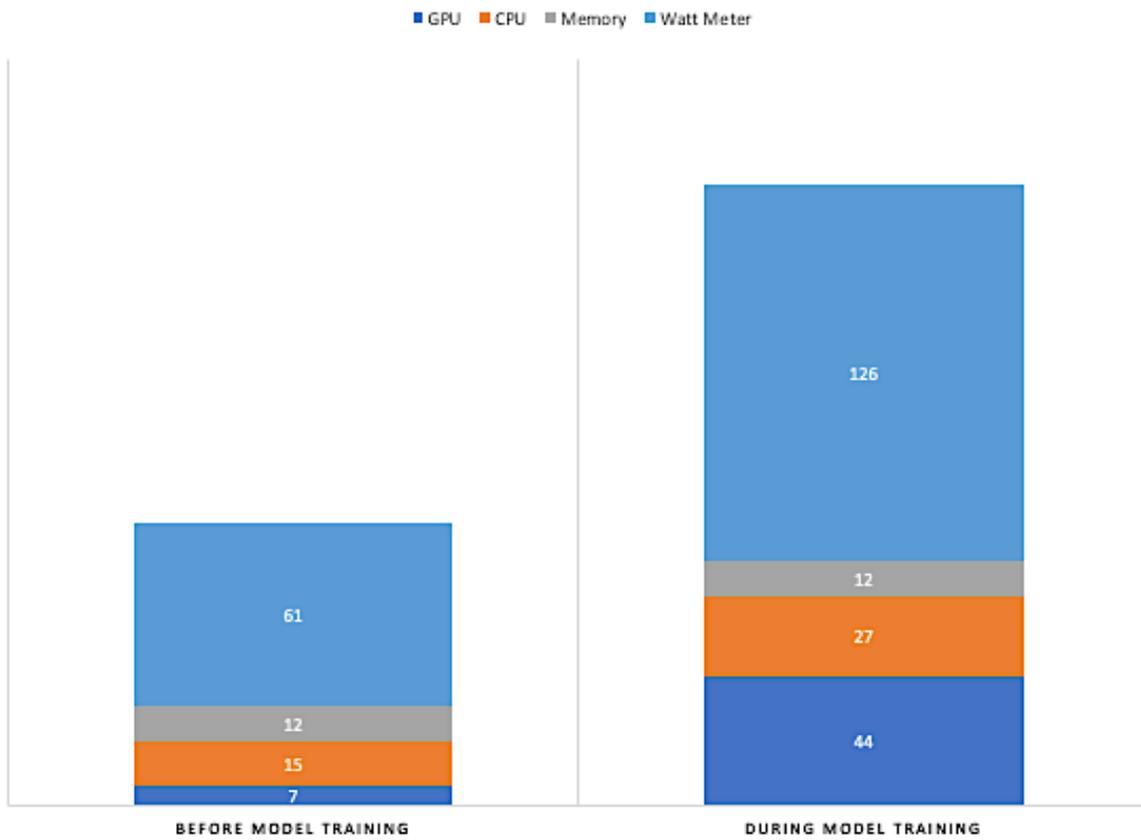

**Figure 9: Power consumption data for the Parquet regression benchmarking**

Figure 10 indicates that the GPU was used during the model training, and the utilisation was 40%, similar to the CSV test.

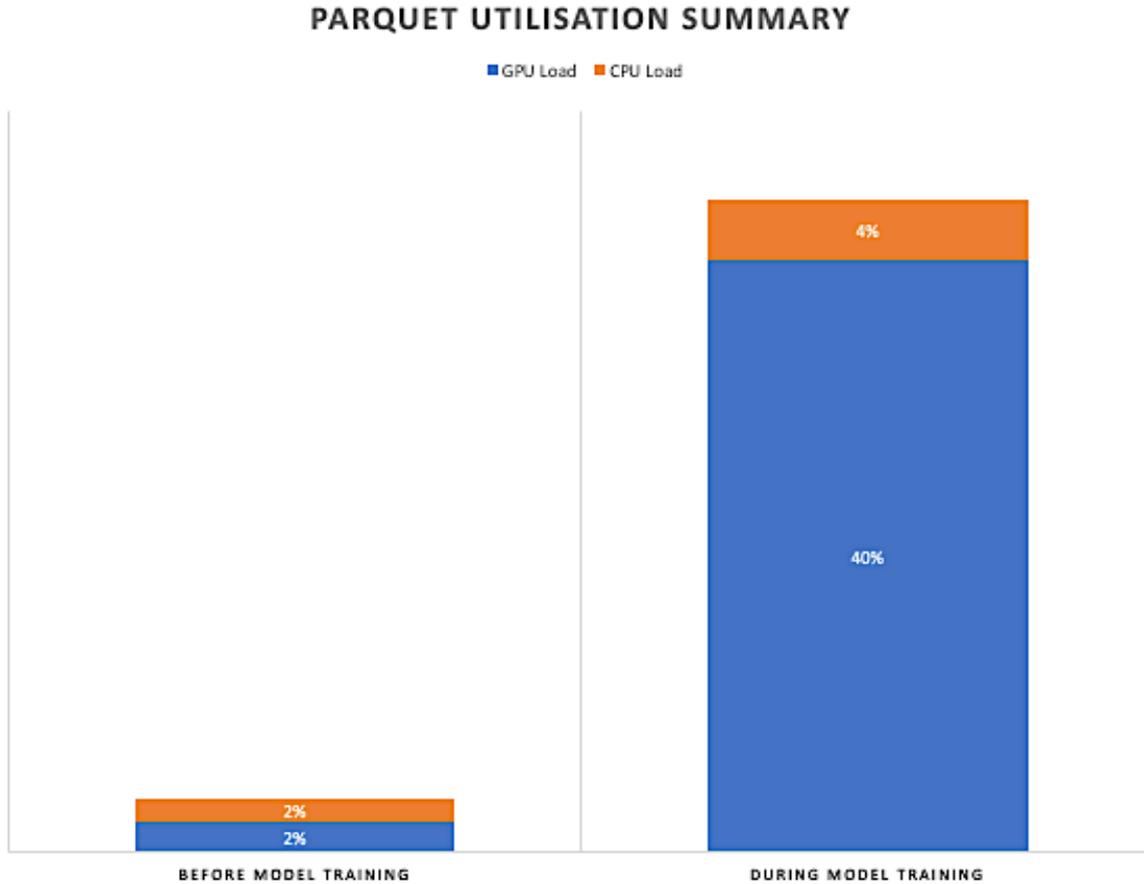

**Figure 10: Hardware utilisation data for the Parquet regression benchmarking**

Overall, we can see that the power consumption and GPU utilisation are the same between the two dataset formats.

### 4.4 Experiments

For each regression experiment, a procedure similar to the classification [1] was followed. Different batches and neurons were chosen to produce a variety of results that could be compared with the benchmarking data. The common factor is the mixed precision and the epochs, which keep the same model training duration between experiments.
Table 3 shows the DNN network configuration for each CSV and Parquet experiment. The CSV experiments used the same configuration as the associated Parquet experiment to provide a fair comparison.

|  | Floating Point | Batch Size | Neurons | Epochs |
| --- | --- | --- | --- | --- |
| CSV – 1st Experiment | Mixed Precision | 256 | 1024 | 1500 |
| CSV – 2nd Experiment | Mixed Precision | 512 | 1024 | 1500 |
| CSV – 3rd Experiment | Mixed Precision | 1024 | 2048 | 1500 |
| Parquet – 1st Experiment | Mixed Precision | 256 | 1024 | 1500 |
| Parquet – 2nd Experiment | Mixed Precision | 512 | 1024 | 1500 |
| Parquet – 3rd Experiment | Mixed Precision | 1024 | 2048 | 1500 |

Table 3: Hyper-parameters for the regression experiments

Before training the model, measurements were taken to validate the PC's status and ensure unnecessary processes were not active. Figure 11 and Figure 12 show the power consumption in Watts of the GPU, CPU, and RAM for both dataset formats. GPUs have low consumption because they are not utilised yet, while CPUs have higher consumption because of the activities within the operating system. The RAM is a fixed number due to the limitation of Code Carbon, as explained earlier.

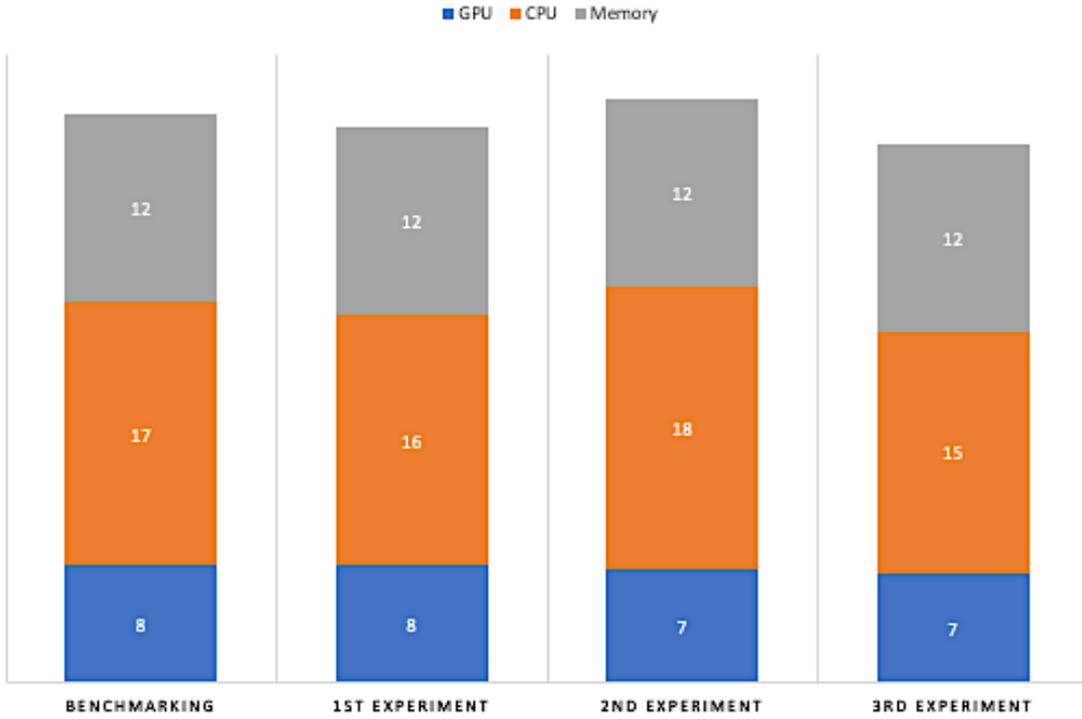

Figure 11: CSV Power consumption data before the model training

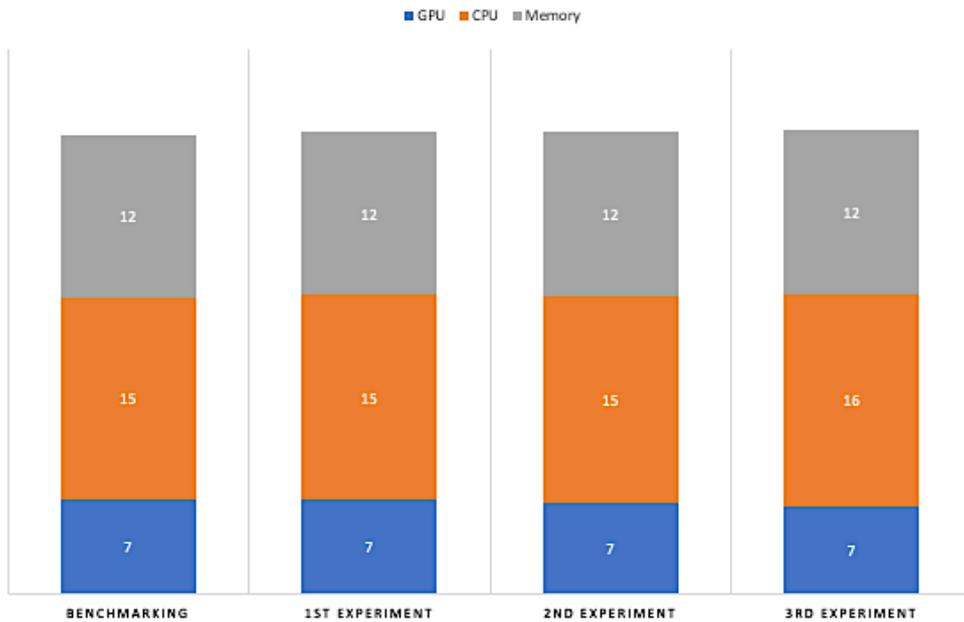

Figure 12: Parquet Power consumption data before the model training

Also, as shown in Figure 13, the utilisation between the GPU and CPU reports normal values because the PC is idle.

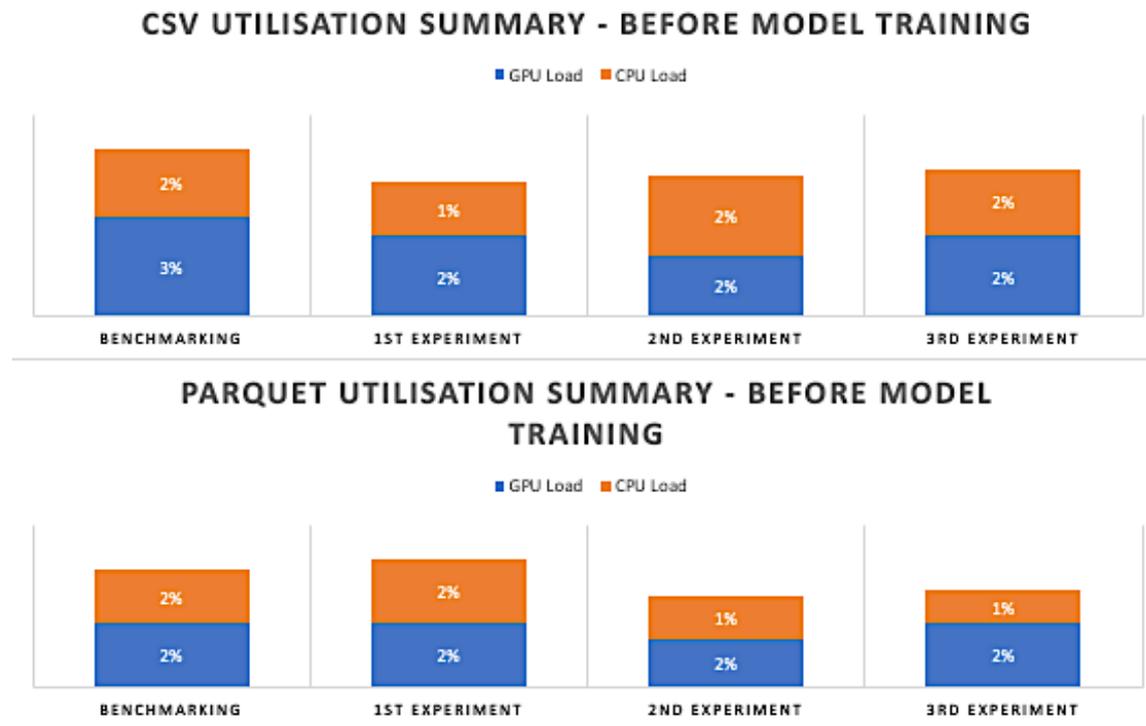

Figure 13: CSV & Parquet hardware utilisation data before model training

Figure 14 shows the overall power consumption in Watts before the model training for both dataset formats.

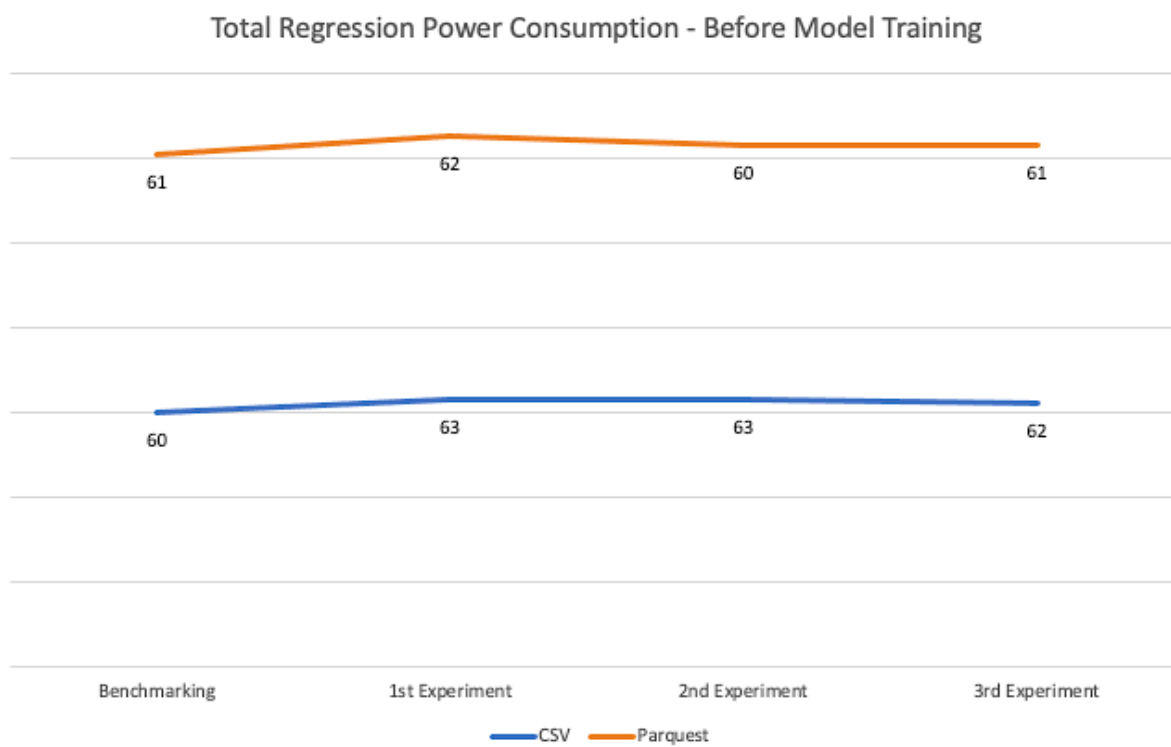

Figure 14: CSV & Parquet total power consumption before model training

During the model training for the CSV dataset, there seems to be a consistency in the GPU power consumption, except in the 2nd experiment, which dropped to 30 watts because it uses 512 batch size instead of 256 but the same number of neurons as per the benchmarking and 1st experiment, as shown in Figure 15.
However, the 3rd experiment reports slightly higher GPU power consumption mainly because of the double amount of neurons, which is 2048 instead of 1024.

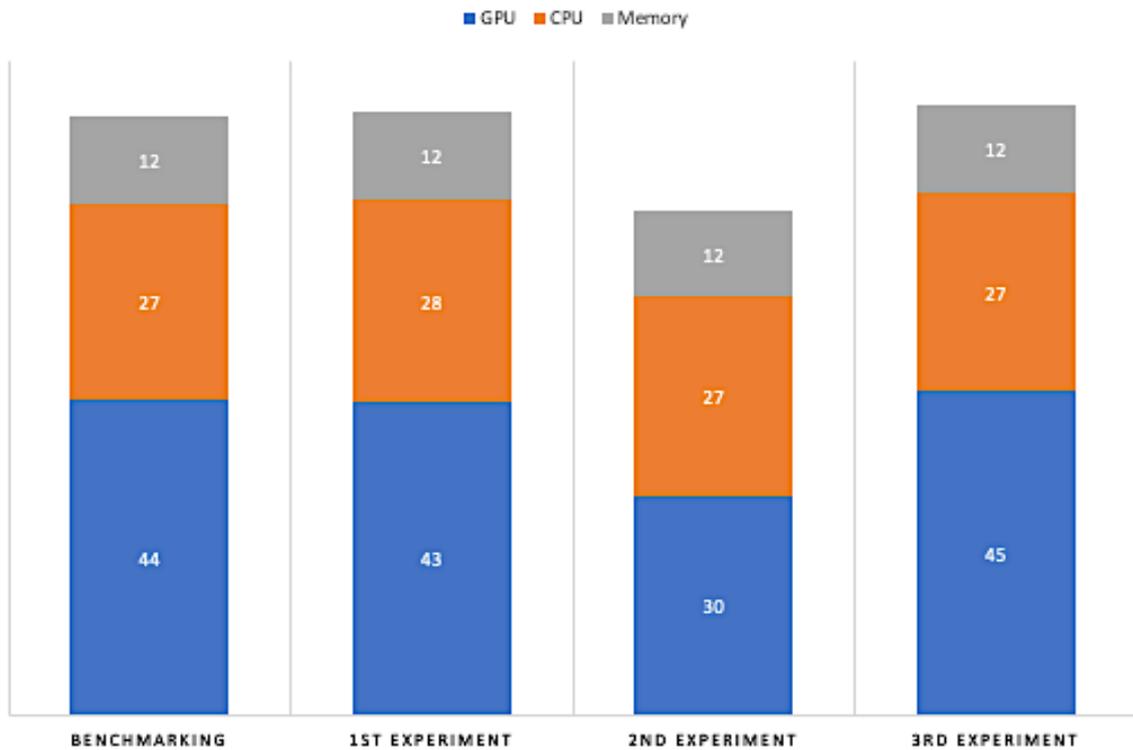

Figure 15: CSV Power consumption data during the model training

A similar pattern appears with the Parquet tests. The 2nd experiment uses less GPU wattage than the other experiments due to the larger batch size, but the 3rd experiment's GPU wattage matches the benchmarking power consumption, as shown in Figure 16.

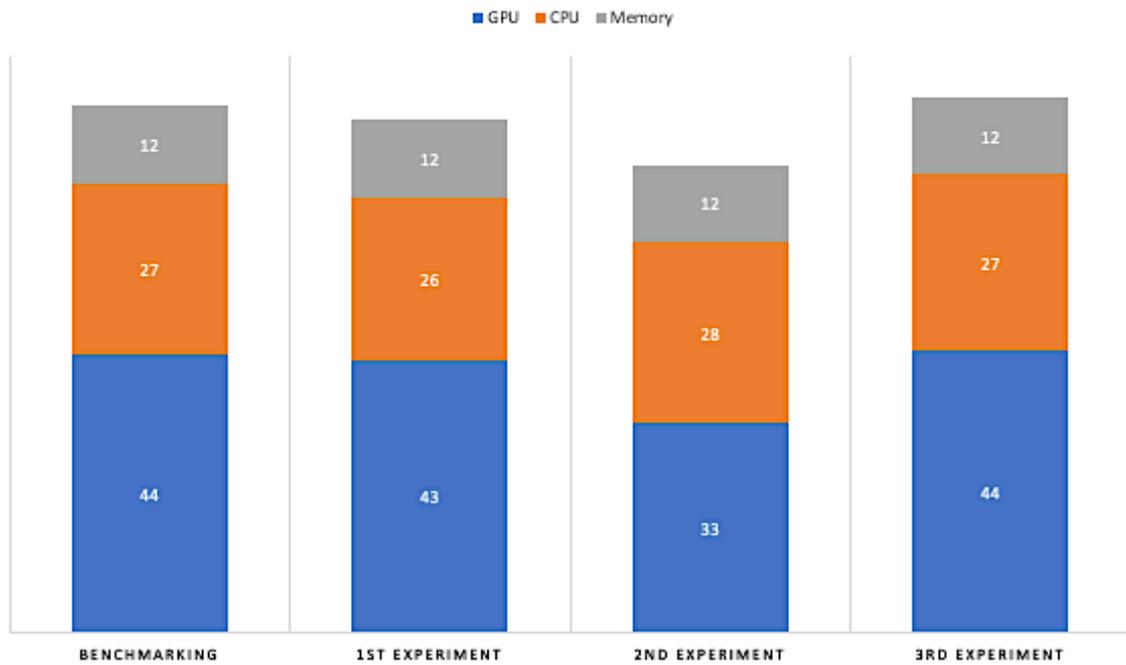

**Figure 16: Parquet Power consumption data during the model training**

Likewise, the GPU usage in both dataset formats was similar, except for the third experiment, which reported higher values because of a higher number of neurons. Figure 17 illustrates the hardware usage and confirms the GPU usage during the model training.

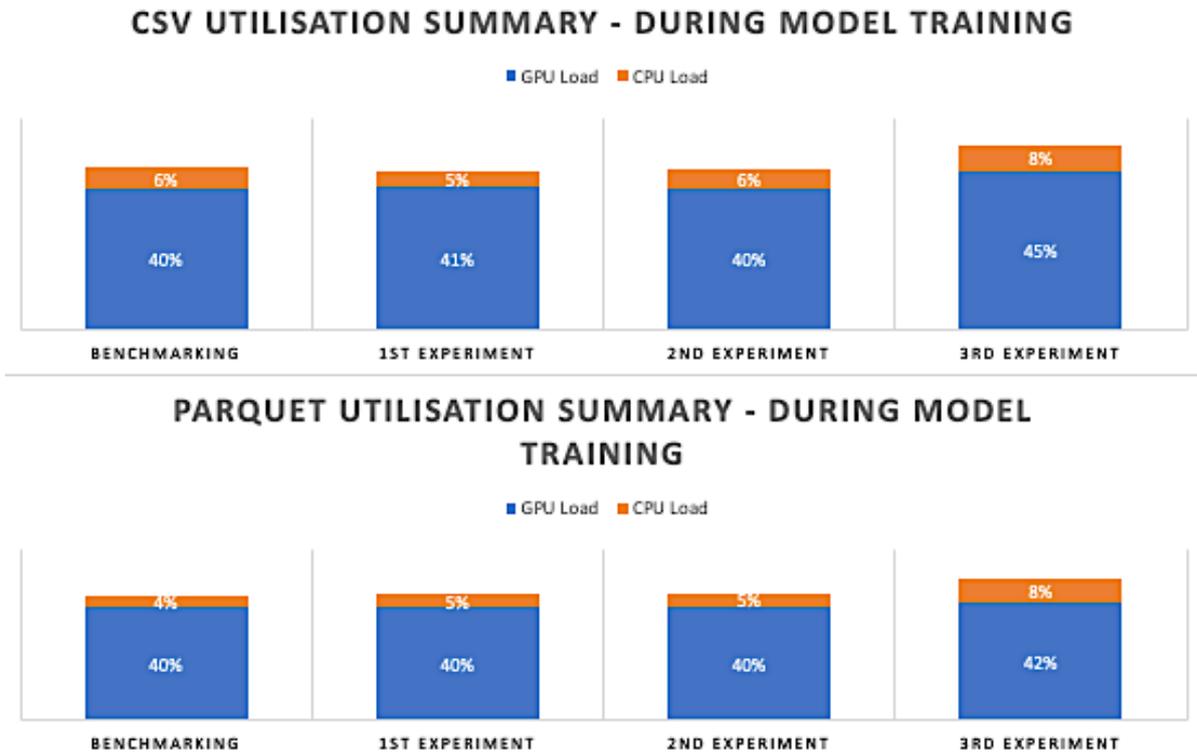

**Figure 17: CSV & Parquet Hardware utilisation data during the model training**

Figure 18 shows the overall power consumption for the CSV and Parquet test and confirms the previous findings. The benchmarking and the 1st experiment have a similar power consumption. However, the 2nd experiment has less power consumption among all. Moreover, the 3rd experiment is the exception because of the higher number of neurons, which forces the GPU to work harder.

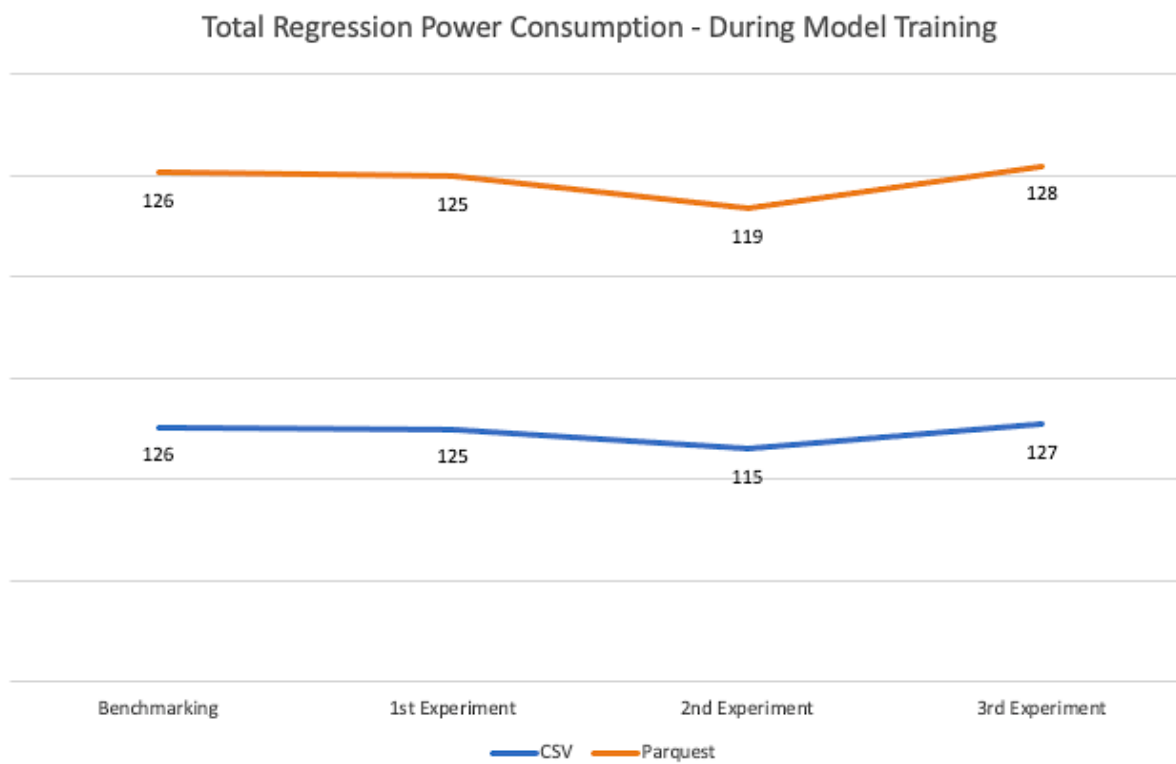

**Figure 18: CSV & Parquet total power consumption data during the model training**

The benchmarking and the 1st experiment provided similar results between the CSV and Parquet. During the 2nd experiment, there was a significant reduction in the GPU and the overall power consumption, which indicates that the bigger batch sizes with mixed precision can help to improve the power consumption. However, as per the 3rd experiment, the power consumption and utilisation also increase when the number of neurons increases.

The same steps as the classification part [1] were followed to calculate the carbon footprint for the regression tests, and the results are shown in Table 4. The only difference is that the model training duration was 1 hour, and the carbon intensity was 123 gCO2e because the tests took place on different days.

The outcome is that the 2nd experiment used a lower carbon footprint mainly because of lower power consumption.

|  | Benchmarking | 1st Experiment | 2nd Experiment | 3rd Experiment |
|---|---|---|---|---|
| CSV Power Consumption | 126 Watts | 125 Watts | 115 Watts | 127 Watts |
| CSV Carbon Footprint | 15.498 gCO2e/kWh | 15.375 gCO2e/kWh | 14.145 gCO2e/kWh | 15.621 gCO2e/kWh |
| Parquet Power Consumption | 126 Watts | 125 Watts | 119 Watts | 128 Watts |
| Parquet Carbon Footprint | 15.498 gCO2e/kWh | 15.375 gCO2e/kWh | 14.637 gCO2e/kWh | 15.744 gCO2e/kWh |

**Table 4: CSV and Parquet carbon footprint calculation**

The regression tests indicate that 512 batch size and 1024 neurons with mixed precision can produce better results. However, as the literature states, hyper-parameters can affect hardware performance; therefore, adjustments will be required to achieve the desired results.

Additionally, there is no difference in using CSV or Parquet dataset format if hyper-parameters are the same. However, when using optimised hyper-parameters with mixed precision, CSV has slightly better power consumption. These results confirm the suggestion from the literature that Parquet is not optimised for GPU-based tasks.

## 5 Analysis and Evaluation

### 5.1 Introduction

During the analysis, the same four groups as per the classification [1] were used to identify a potential statistical significance based on their means using the ANOVA test, as shown in Figure 19. Each group has four values: GPU, CPU, RAM and total power consumption, which were taken from the Wattmeter.

ANOVA can be used when we have more than two groups, but if there is a significant difference, it does not illustrate where the significance lies [27]. Therefore, multiple T-tests have been used to compare the means between a combination of two groups [27].

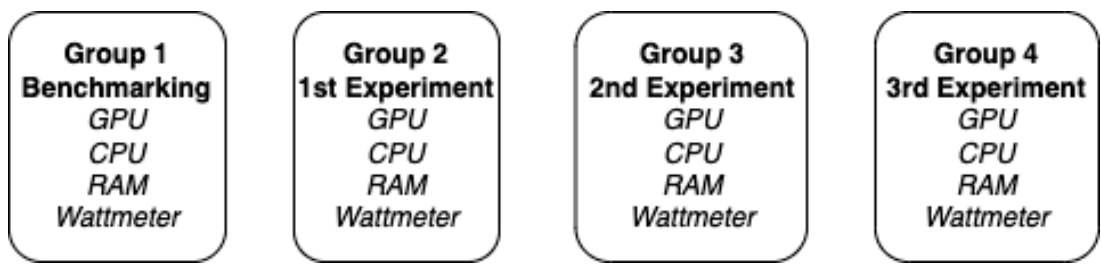

**Figure 19: Groups used for the inferential analysis**

## 5.2 Regression Analysis

Both regression tests, for the CSV and Parquet, have been conducted using the same principles as the classification analysis [1].
The assumptions were the following [31]:
1. The data in each group are normally distributed
2. The data in each group have the same variance
3. The data are independent

Tables 5 and Table 6 summarise the data collected during each regression test for the CSV and Parquet.

|  | Benchmarking<br>Floating Point: 32<br>Batch Size: 256<br>Neurons: 1024<br>Epochs: 1500 | 1st Experiment<br>Floating Point: MP<br>Batch Size: 256<br>Neurons: 1024<br>Epochs: 1500 | 2nd Experiment<br>Floating Point: MP<br>Batch Size: 512<br>Neurons: 1024<br>Epochs: 1500 | 3rd Experiment<br>Floating Point: MP<br>Batch Size: 1024<br>Neurons: 2048<br>Epochs: 1500 |
|---|---|---|---|---|
| GPU | 44 Watts | 43 Watts | 30 Watts | 45 Watts |
| CPU | 27 Watts | 28 Watts | 27 Watts | 27 Watts |
| RAM | 12 Watts | 12 Watts | 12 Watts | 12 Watts |
| Wattmeter | 126 Watts | 125 Watts | 115 Watts | 127 Watts |

**Table 5: CSV Power consumption data during model training**

|  | Benchmarking | 1st Experiment | 2nd Experiment | 3rd Experiment |
|---|---|---|---|---|
|  | Floating Point: 32<br>Batch Size: 256<br>Neurons: 1024<br>Epochs: 1500 | Floating Point: MP<br>Batch Size: 256<br>Neurons: 1024<br>Epochs: 1500 | Floating Point: MP<br>Batch Size: 512<br>Neurons: 1024<br>Epochs: 1500 | Floating Point: MP<br>Batch Size: 1024<br>Neurons: 2048<br>Epochs: 1500 |
| GPU | 44 Watts | 43 Watts | 33 Watts | 44 Watts |
| CPU | 27 Watts | 26 Watts | 28 Watts | 27 Watts |
| RAM | 12 Watts | 12 Watts | 12 Watts | 12 Watts |
| Wattmeter | 126 Watts | 125 Watts | 119 Watts | 128 Watts |

Table 6: Parquet Power consumption data during model training

Data normality was validated using Skewness and Kurtosis. The accepted values for Skewness are between -2 and +2, and the kurtosis value is between -7 and +7 [32]. As shown in Table 7 and Table 8 for CSV and Parquet, the values are within the expected ranges and very close to each other. Therefore, the distribution is normal.

|  | Benchmarking | 1st Experiment | 2nd Experiment | 3rd Experiments |
|---|---|---|---|---|
| Kurtosis | 2.775735592 | 2.790244464 | 3.487471001 | 2.689415071 |
| Skewness | 1.627953799 | 1.623051355 | 1.817803545 | 1.60520868 |

Table 7: CSV Kurtosis and Skewness values between tests

|  | Benchmarking | 1st Experiment | 2nd Experiment | 3rd Experiments |
|---|---|---|---|---|
| Kurtosis | 2.771883958 | 2.815522938 | 3.405256573 | 2.761005135 |
| Skewness | 1.628756088 | 1.649092898 | 1.788594826 | 1.622936199 |

Table 8: Parquet Kurtosis and Skewness values between tests

The F-test was adopted to compare the variance between the tests, and six tests for CSV and Parquet were completed, as shown in Table 9 and Table 10. If the value is closer to 1.5 or less, the sample variance is equal, and we can confidently perform the ANOVA test [33]. All values are close to 1, indicating that the sample variance is within the acceptance range and allowing us to proceed with the ANOVA test. Also, the data are independent because different use cases are being tested between the groups, which validates the third assumption.

|  | Benchmarking | 1st Experiment | 2nd Experiment |
|---|---|---|---|
| 1st Experiment | 1.02397323 |  |  |
| 2nd Experiment | 1.193869873 | 1.165919028 |  |
| 3rd Experiment | 0.985141568 | 0.962077464 | 0.82516662 |

**Table 9: CSV F-Test variance comparison between tests**

|  | Benchmarking | 1st Experiment | 2nd Experiment |
|---|---|---|---|
| 1st Experiment | 1.008965392 |  |  |
| 2nd Experiment | 1.119296002 | 1.109350242 |  |
| 3rd Experiment | 0.968321995 | 0.959717749 | 0.865116996 |

**Table 10: Parquet F-Test variance comparison between tests**

The P-value, 0.05, was used to calculate the ANOVA test for the CSV tests, and the results can be seen in Table 11 and Table 12. The 2nd experiment had better results based on the average. However, the calculated P-value is higher than 0.05, and the F-value is smaller than the F-critical, which indicates that we can accept the $H_0$.

| Groups | Count | Sum | Average (Mean) | Variance |
|---|---|---|---|---|
| Benchmarking | 4 | 208.5 | 52.125 | 2591.0625 |
| 1st Experiment | 4 | 208.05 | 52.0125 | 2530.400625 |
| 2nd Experiment | 4 | 184.65 | 46.1625 | 2170.305625 |
| 3rd Experiment | 4 | 210.95 | 52.7375 | 2630.142292 |

Table 11: CSV ANOVA Single Factor summary for each group

| Source of Variation | SS | Df | MS | F-value | P-value | F-critical |
|---|---|---|---|---|---|---|
| Between Groups | 113.9179688 | 3 | 37.97265625 | 0.015308606 | 0.997269943 | 3.490294819 |
| Within Groups | 29765.73313 | 12 | 2480.47776 | | | |
| Total | 29879.65109 | 15 | | | | |

Table 12: CSV ANOVA Single Factor P-value calculation

The Parquet ANOVA test uses a similar P-value as the CSV test, and the results are shown in Table 13 and Table 14. The mean of the 2nd experiment had better results, identical to the CSV ANOVA. However, the CSV's 2nd experiment had a lower mean among all the tests in the Parquet results.

The P-value in the Parquet ANOVA test was higher than 0.05, and the F-value is smaller than the F-critical, which indicates that we can accept the $H_0$.

| Groups | Count | Sum | Average (Mean) | Variance |
|---|---|---|---|---|
| Benchmarking | 4 | 208.3 | 52.075 | 2594.4225 |
| 1st Experiment | 4 | 205.1 | 51.275 | 2571.369167 |
| 2nd Experiment | 4 | 192.15 | 48.0375 | 2317.905625 |
| 3rd Experiment | 4 | 211.65 | 52.9125 | 2679.297292 |

Table 13: Parquet ANOVA Single Factor summary for each grou

| Source of Variation | SS | Df | MS | F-value | P-value | F-critical |
|---|---|---|---|---|---|---|
| Between Groups | 54.57125 | 3 | 18.19041667 | 0.007159471 | 0.999118842 | 3.490294819 |
| Within Groups | 30488.98375 | 12 | 2540.748646 | | | |
| Total | 30543.555 | 15 | | | | |

**Table 14: Parquet ANOVA Single Factor P-value calculation**

Similarly, paired T-tests used a P-value of 0.0083, as per the classification analysis [1], and Table 15 shows the calculated P-values. Based on the results, all values are higher than the predefined P-value. Therefore, there is no significant difference between the means of the tests, and we can accept the H$_0$.

| | 1st Experiment | 2nd Experiment | 3rd Experiment |
|---|---|---|---|
| Benchmarking | 0.390767559 | 0.097587467 | 0.068242449 |
| 1st Experiment | | 0.088136553 | 0.161021179 |
| 2nd Experiment | | | 0.094684318 |

**Table 15: CSV One-side paired T-Test P-values comparison**

Parquet paired T-tests used the same P-value of 0.0083, as per the CSV, and Table 16 shows the calculated one-sided P-values. Similar to the CSV results, all values exceed the predefined P-value. Therefore, the means in the specific sample size for the Parquet regression tests are not significantly different, which confirms that we accept the H$_0$.

|  | 1st Experiment | 2nd Experiment | 3rd Experiment |
|---|---|---|---|
| Benchmarking | 0.029911193 | 0.129980915 | 0.070068599 |
| 1st Experiment |  | 0.168736783 | 0.038448405 |
| 2nd Experiment |  |  | 0.109816527 |

Table 16: CSV One-side paired T-Test P-values comparison

This chapter analysed the test data and adopted two analysis techniques: ANOVA and T-tests. Firstly, the validation of the ANOVA assumptions was justified using Skewness, Kurtosis, and F-Test for the tests. Secondly, the ANOVA and T-test results were analysed by comparing the predefined and calculated P-values to be able to accept or reject the null hypothesis.

Based on the findings from the regression analysis, it concluded that even though there is a slight numerical difference between the benchmarking and experiment means, statistically, there is no difference. Therefore, the $H_0$ has been accepted across all the regression tests. This means that changing the independent variables, such as ML optimisation techniques and dataset, will not affect the dependent variable or the computing power consumption.

### 5.3 Limitations of the Analysis

The propositions might not be valid because the analysis has limitations. To begin with, the RAM power consumption is based on a fixed value provided by the software. The CPU and GPU measurements varied between the software and required using the average value. However, by monitoring the utilisation, the author could verify the GPU usage across all the tests. Additionally, manually calculating the average for the Wattmeter values was required.

A significant limitation is the sample size, which included a single GPU, CPU, RAM and Wattmeter. Therefore, if there is a slight difference in the relationship between variables or groups, as in this research, a large sample size will help obtain a more accurate statistic test [34].

## 6 Conclusion and Future Work

In this research, the author discussed the potential improvement of the ML carbon footprint by investigating different ML optimisation techniques. Current literature suggests that using mixed precision during the ML model training can improve the GPU's computation and performance [30]. Additionally, it requires using hyper-parameters, which are essential for creating DNN networks [22], but it is important to configure the hyper-parameters appropriately to avoid negatively affecting the GPU performance [24].

As the literature suggested [21], various software must be used to monitor the hardware utilisation and computation power consumption because of existing limitations.

The regression tests used two dataset formats, CSV and Parquet. The author used the benchmarking test as a reference point using default DNN parameters and then completed a series of experiments using the literature suggestions. Initially, the results were compared with the associated benchmarking results using descriptive analysis, specifically the mean. The regression results show that while mixed precision can help improve power consumption, we must carefully consider the hyper-parameters. A high number of batch sizes and neurons will negatively affect power consumption.

After summarising the test results, the author analysed the data using inferential statistics, specifically ANOVA and T-test. The commonality between the two tests is that both compare the means between groups. Still, ANOVA has a different way of determining the statistical significance and can be used when there are more than three groups. Therefore, the author used ANOVA to compare the benchmarking with the three experiments. After the ANOVA comparison, the author used a T-test to compare multiple pairs of groups to cross-validate the ANOVA results and reduce Type I and Type II errors. The results reported no statistical significance between the means in the regression test and accepted $H_0$. Therefore, choosing different ML techniques and the Parquet dataset format will not improve the computational power consumption and the overall ML carbon footprint.

However, some limitations can affect the generalisation of this research, which future studies can probably overcome. First, future research could use a single software that supports a wide range of hardware to collect power consumption data more accurately and frequently. Second, future research may use a larger implementation with a cluster of GPUs, which will help to increase the sample size significantly because it is an essential factor for statistical analysis and can affect the outcome [34].

[21] Eco2AI: carbon emissions tracking of machine learning models as the first step towards sustainable AI, August 2023. [Online]. Available:
http://arxiv.org/abs/2208.00406

[22] Hyper-parameter Optimization of a Convolutional Neural Network, November 2023. [Online]. Available:
https://scholar.afit.edu/cgi/viewcontent.cgi?params=/context/etd/article/3298/&path_info=AFIT_ENS_MS_19_M_105_Chon_S.pdf

[23] Hyperparameter Tuning, 2029. [Online]. Available:
http://rgdoi.net/10.13140/RG.2.2.11820.21128

[24] Assessing Hyper Parameter Optimization and Speedup for Convolutional Neural Networks, July 2020. [Online]. Available: http://services.igi-global.com/resolvedoi/resolve.aspx?doi=10.4018/IJAIML.2020070101

[25] Determining the Number of Neurons in Artificial Neural Networks for Approximation, Trained with Algorithms Using the Jacobi Matrix, November 2020. [Online]. Available:
http://www.temjournal.com/content/94/TEMJournalNovember2020_1320_1329.html

[26] The effect of batch size on the generalizability of the convolutional neural networks on a histopathology dataset, December 2020. [Online]. Available:
https://linkinghub.elsevier.com/retrieve/pii/S2405959519303455

[27] Quantitative Data Analysis, May 2021. [Online]. Available:
http://rgdoi.net/10.13140/RG.2.2.23322.36807

[28] A Statistical Primer: Understanding Descriptive and Inferential Statistics, March 2007. [Online]. Available:
https://journals.library.ualberta.ca/eblip/index.php/EBLIP/article/view/168

[29] Honghui Zhou, Ruyi Qin, Zihan Liu, Ying Qian, and Xiaoming Ju "Optimizing Performance of Image Processing Algorithms on GPUs
" in Proceeding of 2021 International Conference on Wireless Communications, Networking and Applications, 2022, pp. 936-942

[30] Mixed Precision Training, February 2018. [Online]. Available:
http://arxiv.org/abs/1710.0374034